\documentclass{article}

    \usepackage[final]{neurips_2021_ml4ad}

\usepackage[utf8]{inputenc} % allow utf-8 input
\usepackage[T1]{fontenc}    % use 8-bit T1 fonts
\usepackage{hyperref}       % hyperlinks
\usepackage{url}            % simple URL typesetting
\usepackage{booktabs}       % professional-quality tables
\usepackage{amsfonts}       % blackboard math symbols
\usepackage{nicefrac}       % compact symbols for 1/2, etc.
\usepackage{microtype}      % microtypography
\usepackage{xcolor}         % colors
\usepackage{amssymb}        % checkmark
\usepackage{pifont}         % crossmark
\usepackage{graphicx}
\usepackage{babel}
\usepackage{caption}
\usepackage[frozencache=true,cachedir=minted-cache]{minted}
\usepackage{enumitem}
\usepackage{makecell}
\usepackage{amsmath}
\usepackage{hyperref}
\usepackage{multirow}

\newcommand{\xmark}{\ding{55}}

\hypersetup{
    colorlinks=true,
    linkcolor=blue}

\definecolor{ego_red}{RGB}{166,61,55}
\definecolor{agent_blue}{RGB}{58,119,175}
\definecolor{trajectory_green}{RGB}{60,109,33}
\definecolor{purple_ours}{RGB}{102,78,188}

\graphicspath{ {./figures/} }
\usepackage{natbib}
\title{DriverGym: Democratising Reinforcement Learning for Autonomous Driving}

\author{%
  Parth Kothari\\
  Level 5, Woven Planet\\
   \And
   Christian Perone \\
   Level 5, Woven Planet\\
    \And
   Luca Bergamini \\
   Level 5, Woven Planet\\
  
    \And
   Alexandre Alahi \\
   EPFL \\
  
    \And
   Peter Ondruska \\
   Level 5, Woven Planet\\
}

\begin{document}

\maketitle

\begin{abstract}
 Despite promising progress in reinforcement learning (RL), developing algorithms for autonomous driving (AD) remains challenging: one of the critical issues being the absence of an open-source platform capable of training and effectively validating the RL policies on real-world data. We propose DriverGym, an open-source OpenAI Gym-compatible environment specifically tailored for developing RL algorithms for autonomous driving. DriverGym provides access to more than 1000 hours of expert logged data and also supports reactive and data-driven agent behavior. The performance of an RL policy can be easily validated on real-world data using our extensive and flexible closed-loop evaluation protocol. In this work, we also provide behavior cloning baselines using supervised learning and RL, trained in DriverGym. Code and videos are available on the \href{https://lyft.github.io/l5kit/}{L5Kit repository}.
%  We make DriverGym code, as well as all the baselines publicly available to further stimulate development from the community. 
\end{abstract}

\begin{figure}[h!]
    \centering
    \includegraphics[width=0.95\textwidth]{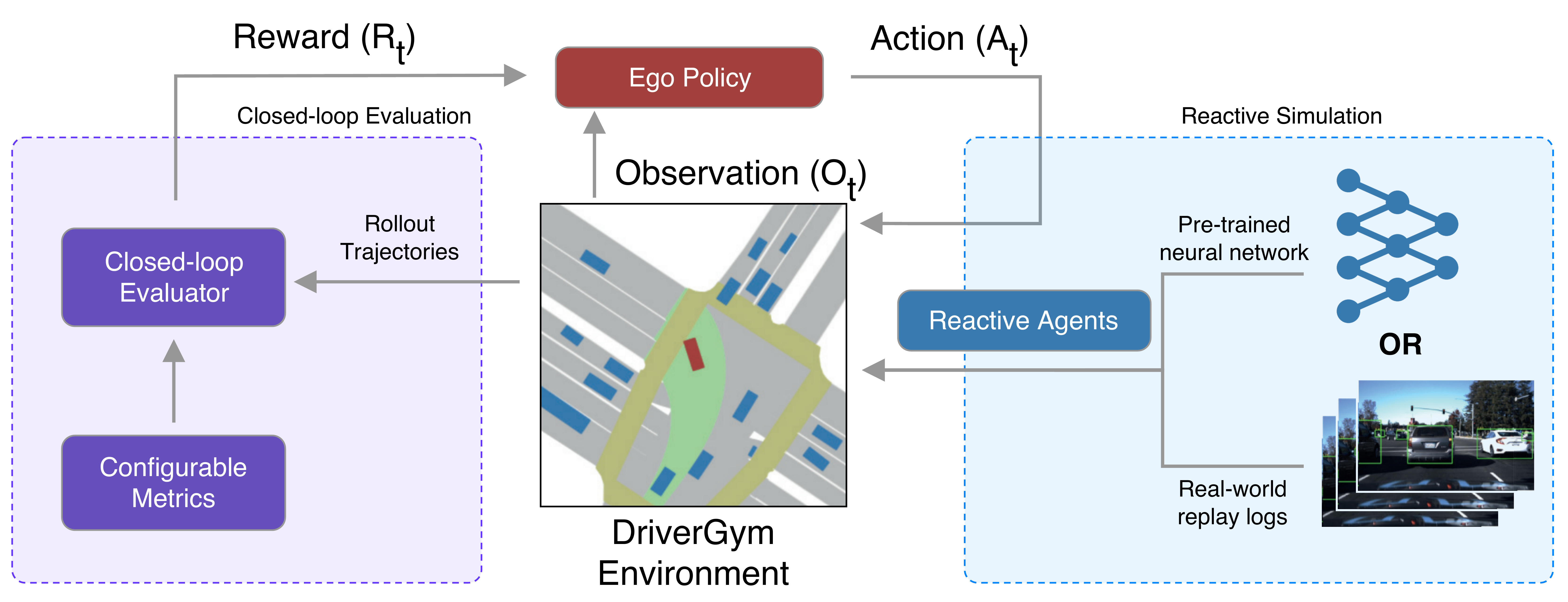}
    \caption{DriverGym: an open-source gym environment that enables training RL driving policies on real-world data. The RL policy can access rich semantic maps to control the ego (\textcolor{ego_red}{red}). Other agents (\textcolor{agent_blue}{blue}) can either be simulated from the data logs or controlled using a dedicated policy pre-trained on real-world data. We provide an extensible evaluation system (\textcolor{purple_ours}{purple}) with easily configurable metrics to evaluate the idiosyncrasies of the trained policies.}
    \label{fig:pullfig}
\end{figure}

\section{Introduction}

Recently, Reinforcement Learning (RL) has achieved great success in a variety of applications like playing Atari games~\citep{Mnih2013PlayingAW}, board games~\citep{Silver2017MasteringCA}, manipulating sensorimotor in three-dimensions~\citep{MartnMartn2019VariableIC}. However, developing RL algorithms for real-world applications such as autonomous driving (AD) remains an open challenge~\citep{Kiran2020DeepRL}: with AD being an extremely safety-critical task, one cannot directly deploy a policy in the real world for data collection or policy validation.

One solution is to deploy the policy in the real world with a safety driver inside the car at all times. However, this process is time-consuming, and more importantly, not accessible to all of the research community. Therefore, to tackle this challenge, there is a dire need for an RL simulation environment that can (1) be used to easily train RL policies using real-world logs, (2) simulate surrounding agent behavior that is both realistic and reactive to the ego policy, (3) effectively evaluate the trained models, (4) be flexible in its design, and (5) inclusive to the entire research community.
% to data collection or model validation

% Many simulation environments have been proposed in recent times to foster the development of autonomous driving policies, the most notable platform being CARLA (\cite{Dosovitskiy2017CARLAAO}). CARLA supports flexible specification of sensor suites and environmental conditions for training and testing. However, being a synthetic simulator, CARLA does not provide access to real-world driving data. Further, the lack of gym compatibility limits the direct applicability of various RL policies.

We propose DriverGym, an open-source gym-compatible environment specifically tailored for developing and experimenting with RL algorithms for self-driving (see Fig.~\ref{fig:pullfig}). DriverGym utilizes one of the largest public self-driving datasets, \textit{Level 5 Prediction Dataset}~\citep{Woven} containing over 1,000 hours of data, and provides support for reactive agent behavior simulation~\citep{Bergamini2021SimNetLR} using data-driven models. Furthermore, DriverGym provides an extensive and extensible closed-loop evaluation system: it not only comprises a variety of AD-specific metrics but also can be easily extended to incorporate new metrics to evaluate idiosyncrasies of trained policies. We open-source the code and pre-trained models to stimulate development.

% Our contributions are summarized below:
% \begin{itemize}[leftmargin=*]
% \itemsep0em 
% \item A Gym-compatible environment that (1) uses more than 1000 hours of real-world data (2) enables reactive and realistic agent simulations (3) uses an extensive and extensible evaluation protocol;
% \item The environment, pre-trained models and behavior cloning scripts using supervised learning and reinforcement learning are open-source to democratize research in autonomous driving.
% \end{itemize}
\newcolumntype{?}{!{\vrule width 1pt}}
\renewcommand\theadalign{bc}
\renewcommand\theadfont{\bfseries}
\renewcommand\theadgape{\Gape[4pt]}
\renewcommand\cellgape{\Gape[4pt]}

\begin{table}
  \centering
  \caption{Comparison of various open-source RL simulation environments for autonomous driving.}
%   DriverGym allows simulating agents using data-driven models as well as using real-world log data.
  \label{rl-table}
  \centering
%   \begin{tabular}{p{2.7cm}p{1.6cm}p{2.4cm}p{2.0cm}p{1.4cm}p{1.4cm}}
  \resizebox{\textwidth}{!}{\begin{tabular}{lccccc}
    \toprule
    Name & \makecell{Gym \\ Compatible} & \makecell{Evaluation \\ Protocol} & \makecell{Simulator \\ Expert Data} & \makecell{Real-world \\ Expert Data}  & \makecell{Agents \\ Model} \\
    \midrule
    TORCS & \xmark & \xmark & \xmark & \xmark & Rule-based\\
    Highway-Env  & \checkmark & \xmark & \xmark & \xmark  & Rule-based  \\
    CARLA (Official)  & \xmark & \checkmark  & 14 hrs & \xmark  & Rule-based\\
    % DeepDrive  & \xmark& \xmark & 8.2 hrs & \xmark  & Rule-based \\
    SMARTS & \checkmark & \checkmark & \xmark & \xmark  & Rule-based \textit{or} Data-driven\\
    CRTS & \checkmark & \checkmark & \xmark & 64 hrs & Real-world Logs\\
    DriverGym & \checkmark & \checkmark & \xmark & 1000 hrs  & Data-driven \textit{or} Real-world Logs\\
    \bottomrule
  \end{tabular}}
\end{table}

In this work, we provide the following contributions:
%In summary, our open-source gym-compatible environment DriverGym has the following features:
\begin{itemize}
\itemsep0em 
\item An open-source and OpenAI gym-compatible environment for autonomous driving task;
\item Support for more than 1000 hours of real-world expert data;
% \item Enables reactive and realistic agent trajectory simulations;
\item Support for logged agents replay or data-driven realistic agent trajectory simulations;
\item Configurable and extensible evaluation protocol;
\item Provide pre-trained models and the corresponding reproducible training code.
\end{itemize}

% This is because the research community is hindered by the absence of an open-sourced infrastructure capable of training and testing the RL policies on real-world data.

\section{Related Work}

% \subsection{Autonomous Driving Simulation Environments}

To replicate the success of the OpenAI Gym framework~\citep{Brockman2016OpenAIG}, many simulation environments have been developed in the context of autonomous driving~\citep{Espi2005TORCSTO, highway, Dosovitskiy2017CARLAAO, SUMO2018, deepdrive}. Table~\ref{rl-table} provides a comparison amongst commonly used RL simulation environments including DriverGym. Racing simulators like TORCS~\citep{Espi2005TORCSTO} offer limited scenarios of driving. Highway-Env~\citep{highway} provides a collection of gym-compatible environments for autonomous driving. However, it lacks important semantic elements like traffic lights, an extensive evaluation protocol and expert data.

Traffic simulators like CARLA~\citep{Dosovitskiy2017CARLAAO}, SUMO~\citep{SUMO2018} supports flexible specification of traffic conditions for training and testing. However, they are synthetic simulators that utilize hand-coded rules for surrounding agents' motion that tends to be unrealistic and display a limited variety of behaviors. Crucially, they lack access to real-world data logs. SMARTS~\citep{Zhou2020SMARTSSM} overcomes the former issue by providing \textit{Social Agent Zoo} that supports data-driven agent models while CRTS~\citep{Osinski2020CARLART} tackles the latter providing access to 64 hours of real-world logs within the CARLA simulator. DriverGym solves both these challenges: it enables simulating reactive agents using data-driven models learned from real-world data, and provides access to 1000 hours of real-world logs to initialize episodes or simulate agents.

% Moreover, with simulators like CARLA, DeepDrive being based on the Unreal Game Engine, problems like non-determinism and timing issues are introduced~\citep{Chance2021OnDO}, that are undesirable when validating safety-critical RL models.\sergey{how does DriverGym compares to them? we should add a sentence here that it allows using real data but is not reactive. non-determinism and timing issues are not really related to the main contribution, it does not address them directly.}

% \subsection{Datasets for Autonomous Driving}
% To successfully train RL policies, one needs detailed information about the environment in the observation space like semantic maps that encode possible driving behavior to reason about future behaviors. Deep learning solutions leveraging birds-eye-view (BEV) representations of scenes~\citep{Alahi2016SocialLH, Lee2017DESIREDF} or graph neural networks~\citep{Casas2020SpAGNNSG, Gao2020VectorNetEH} have established themselves as the leading solutions for this task.

% We interface DriverGym with the Level 5 Prediction Dataset~\citep{Woven}, comprising 170,000 scenes, to initialize, and optionally train on, real-world scenarios. The dataset contains a high-definition semantic map with 15,242 labeled elements and a high-definition aerial view that equips DriverGym with a rich observation space.

% where each scene is 25 seconds long and captures the perception output of the self-driving system, which encodes the precise positions and motions of nearby vehicles, cyclists, and pedestrians over time

% TODO: christian stopped here, to continue

\section{DriverGym}
% DriverGym has been built to foster the development of RL policies for self-driving by providing a simple and flexible interface to train and evaluate RL policies. 
% Our code is open-source with Apache 2 license.
% Our environment is compatible with both Stable Baselines3~\citep{stable-baselines3} and RLlib~\citep{Liang2018RLlibAF}, two popular frameworks for training state-of-the-art RL policies.
DriverGym aims to foster the development of RL policies for self-driving by providing a flexible interface to train and evaluate RL policies. Our environment is compatible with both SB3~\citep{stable-baselines3} and RLlib~\citep{Liang2018RLlibAF}, two popular frameworks for training RL policies. Our code is open-source with Apache 2 license. We describe the components of our environment below.

% DriverGym has been built to foster the development of RL policies for self-driving by providing a simple and flexible interface to train and evaluate RL policies. Our environment is compatible with both Stable Baselines3~\citep{stable-baselines3} and RLlib~\citep{Liang2018RLlibAF}, two popular frameworks for training state-of-the-art RL policies. Our code is open-source with Apache 2 license.
% 

% thanks to its integrated interface to large-scale real-world data and flexibility to use reactive agents trained in a data-driven manner \textit{e.g.} SimNet model~\citep{Bergamini2021SimNetLR}. 
% It is implemented on top of the OpenAI Gym framework, enabling easy testing of RL policies (even those developed for different applications) on DriverGym. 

% \begin{figure}
%     \centering
%     \includegraphics[width=0.19\textwidth]{S1_F130.png}
%     \hfill
%     \includegraphics[width=0.19\textwidth]{S1_F150.png}
%     \hfill
%     \includegraphics[width=0.19\textwidth]{S1_F170.png}
%     \hfill
%     \includegraphics[width=0.19\textwidth]{S1_F190.png}
%     \hfill
%     \includegraphics[width=0.19\textwidth]{S1_F210.png}
%     \caption{This is the placeholder for episode rollout}
%     \label{fig:rollout}
% \end{figure}

\begin{figure}
    \centering
    \includegraphics[width=\textwidth]{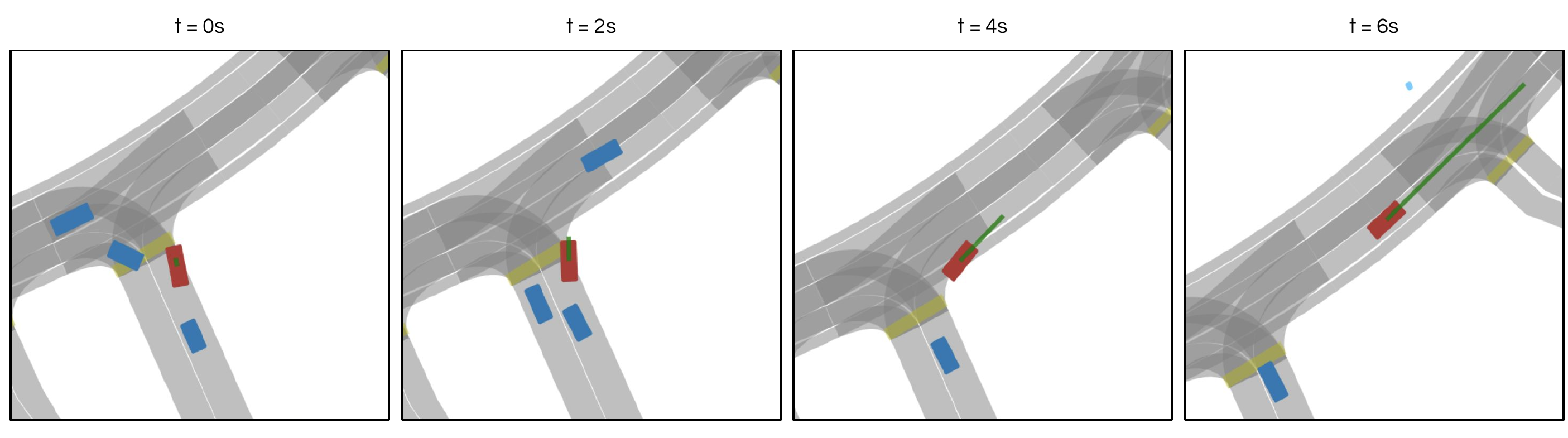}
    % \caption{Visualization of an episode rollout in DriverGym. The policy prediction (green) of ego (red) is scaled by factor of 10 and shown at every 20 frames (10Hz frequency) for better viewing. The agents (blue) are simulated according to real-world data logs.}
    \caption{Visualization of an episode rollout (ego in \textcolor{ego_red}{red}, agents in \textcolor{agent_blue}{blue}) in DriverGym. The policy prediction (\textcolor{trajectory_green}{green} line) is scaled by factor of 10 and shown at 2 second intervals for better viewing.}
    \label{fig:rollout}
\end{figure}

% DriverGym simulates episodes (25 second long sequence) initialized from real world logged data and provides a simple interface between the world and an ego that interacts with the world. The surrounding agents are replayed according to the real-world logged data or simulated using a data-driven approach \cite{Bergamini2021SimNetLR}. 

\subsection{State Representation}
The state representation captures the context around the ego agent, in particular, the surrounding agents' positions, their velocities, the lanes and traffic lights. We encode this information in the form of a 3D tensor that is the birds-eye-view (BEV) raster image of the current frame. DriverGym supports all the rasterization modes provided by L5Kit~\citep{Woven} (see Fig~\ref{raster_viz} in Appendix). Compared to Atari environments, DriverGym requires more time to generate observations as the latter has to load real-world data and subsequently render high-resolution raster images.

% , that create a meaningful semantic representation of the frame (see Fig~\ref{raster_viz} in Appendix). 

\subsection{Action Spaces}
The action produced by the RL policy is used to control the motion of the ego agent. The action is propagated as $(x,y,yaw)$ to update the state of the ego. Still, DriverGym does not make any strict assumptions on the policy itself which can, for instance, output $(acceleration, steer)$ and use a kinematic model to decode the next-step observation.

% DriverGym provides the option of utilizing two  action spaces. In the former, the policy can directly outputs the next-step $(x, y, yaw)$ for the ego agent. In the latter, the policy can output the $(acceleration, steer)$ that is subsequently fed into a kinematic model in DriverGym to calculate the $(x, y, yaw)$. The prediction outputs are used to take a step in the environment and generate the next-step observation.

% The action is propagated to the underneath data log as $(x,y,yaw)$ of the updated ego. Still, DriverGym does not make any strict assumptions on the policy itself, which can for example output $(acceleration, steer)$ and use a kinematic model  to decode the next-step observation in the format that the data suppor

% As mentioned earlier, the surrounding agents move according to the replay logs or a user-defined model (provided to the gym environment).

\subsection{Reactive Agents}
An important component of the DriverGym environment is to model the motion of the surrounding agents. DriverGym allows flexibility in this aspect and currently support two ways of controlling the behavior of surrounding agents: \textit{log replay} and \textit{reactive simulation}. 
% Current RL simulation environments~\citep{Dosovitskiy2017CARLAAO, deepdrive} make agents reactive by scripting them to follow certain rules. However, the hand-coded agents tend to be unrealistic as they rarely present a wide enough variety of behaviors.

In \textit{log replay}, during an episode rollout, the movement of surrounding agents around the ego is replayed in the exact same manner as it happened when the log was collected in the real world. In \textit{reactive simulation}, the agent behavior is both reactive and realistic. Motivated by~\citep{Bergamini2021SimNetLR}, DriverGym allows simulating agent reactivity using data-driven models that learn agent behavior from real-world data, \textit{i.e.,} users can provide neural-network-based agent models trained on real-world data, to simulate agent behavior.

\subsection{Rewards}
The rewards in the environment quantify the performance of a driving policy during a rollout and subsequently guide the training of the policy using reinforcement learning. DriverGym, through the Closed-Loop Evaluation (CLE), supports a variety of AD-specific metrics that are computed per-frame, and can be combined to construct the reward function. This system is described in the section below.

\subsection{Closed-Loop Evaluation Protocol}
Having an extensive closed-loop evaluation (CLE) protocol is a necessity to correctly assess the performance of RL policies before deployment in safety-critical real-world scenarios. Our CLE framework comprises insightful AD-specific metrics: the first set of metrics, specific to imitation learning, are distance-based metrics. % These metrics measure different distance measures between the predicted trajectory of ego and ground-truth trajectory of ego in replay logs. They can be used in cases where we want to include imitation loss-based components in the reward. 
The second set of metrics, specific to safety, capture the various types of collisions that occur between ego and surrounding agents. These include front collision, side collision and rear collisions. More importantly,  our CLE framework can be easily extended to incorporate new metrics that can help to test various properties of the trained policy. 
% The third set of metrics, specific to comfort, account for the sudden changes in acceleration and steer. 
% and ego passiveness~\citep{Bergamini2021SimNetLR}.
An in-depth description of our CLE and its flexibility is provided in the appendix section.

% The various metrics within the CLE framework not only help to guide and regularize the training of RL policies (through reward calculation) but also form the crucial component of the evaluation protocol.

%  \begin{minipage}{\textwidth}
%   \begin{minipage}[b]{0.32\textwidth}
%     \centering
%     \includegraphics[width=\textwidth]{RasterS6.png}
%     \captionof{figure}{Raster Visualization}
%     \label{raster_viz}
 
%   \end{minipage}
%   \hfill
%   \begin{minipage}[b]{0.66\textwidth}
%     \centering
%     \begin{tabular}{lccc}\hline
%       Planning Metric & L2 & L2 + P & PPO \\\hline
%       Front collisions & 10 & 8 & \textbf{2} \\
%       Side collisions & 15 & \textbf{7} & 8  \\
%       Rear collisions & 15 & \textbf{7} & 19  \\
%       Displacement error & 13 & \textbf{5} & 9  \\
%       Distance to reference & 26 & \textbf{10} & 13  \\ \hline
%       ADE & 34.32 & \textbf{12.25} & 17.72  \\
%       FDE & 78.43 & \textbf{22.20} & 42.25  \\\hline
%       \end{tabular}
%       \captionof{table}{Quantitative comparison of three methods using the closed-loop evaluation protocol of DriverGym. Lower is better. The three methods: Supervised learning (SL) using L2 imitation loss [L2], SL with L2 loss plus pertubations [L2 + P], RL using PPO [PPO].}
%       \label{compare}
%     \end{minipage}
%   \end{minipage}

% \newcolumntype{P}[1]{>{\centering\arraybackslash}p{#1}}

\begin{table}[]
     \caption{Evaluation of different training strategies using the CLE protocol in DriverGym. Lower is better. \textbf{SL}: Supervised learning using L2 imitation loss, \textbf{SL + P}: SL plus trajectory pertubations, \textbf{PPO}: RL using PPO, with imitation-loss based reward. More information about metrics and validators in Appendix \ref{sec:cle}. Metrics and validators are in the format: average (std. deviation).} 

    %  Metrics: Average L2 Displacement Error (ADE), Final L2 Displacement Error (FDE). \\ Validators: L2 Displacement Error (L2), Distance to Reference Trajectory (Dist2Ref), Front Collisions, Side Collisions and Rear Collisions.}
     \label{compare}
     \centering
     \bgroup
     \def\arraystretch{1.2}
    % \resizebox{\textwidth}{!}{\begin{tabular}[b]{?P{2cm}?P{2cm}|P{2cm}?P{2cm}|P{2cm}|P{2cm}|P{2cm}|P{2cm}?}
    \resizebox{\textwidth}{!}{\begin{tabular}[b]{? c ? c | c ? c | c | c | c | c ?}

    \hline
        % \multicolumn{1}{c|}{\multirow{3}{*}{Method}} & \multicolumn{2}{c|}{Metrics}  & \multicolumn{5}{c}{Validators}  \\ \cline{2-8}
        % & \multicolumn{2}{c|}{Distance}  &  \multicolumn{2}{c|}{Distance}  &
        %  \multicolumn{3}{c}{Safety} \\\cline{2-8}
        \multicolumn{1}{?c?}{\multirow{2}{*}{\textbf{Method}}} & \multicolumn{2}{c?}{\textbf{Metrics}}  & \multicolumn{5}{c?}{\textbf{Validators}}  \\ \cline{2-8}
        & \makecell{Average \\ Displacement} & \makecell{Final \\ Displacement}  &  \makecell{Final \\ Displacement  \\ (>= 30.0m)} & \makecell{Distance To \\ Reference \\ (>= 4.0m)} & \makecell{Front \\ Collision} & \makecell{Side \\ Collision} & \makecell{Rear \\ Collision} \\ \hline
       SL     & 32.4 $\pm$ 2.7 & 74.5 $\pm$ 5.5 & 9.5 $\pm$ 4.9 & 26 $\pm$ 0.0 & 12.5 $\pm$ 3.5 & 19 $\pm$ 5.6 & 16 $\pm$ 1.4 \\
       SL + P & \textbf{13.4 $\pm$ 1.4} & \textbf{25.5 $\pm$ 3.5} & 5.3 $\pm$ 0.6 & \textbf{9.7 $\pm$ 1.5} & 9 $\pm$ 2.6 & 12.3 $\pm$ 6.8 & \textbf{7 $\pm$ 0.0} \\
       PPO    & 18.7 $\pm$ 2.3 & 46.4 $\pm$ 7.7 & \textbf{4.0 $\pm$ 2.0} & 12.7 $\pm$ 3.2 & \textbf{4.3 $\pm$ 2.5} & \textbf{6 $\pm$ 3.5} & 27 $\pm$ 5.0 \\
    %   PPO + Sim    & 23.8 $\pm$ 2.2 & 60.4 $\pm$ 6.2 & 16.6 $\pm$ 2.2 & 12 $\pm$ 2.6 & 10.6 $\pm$ 4 & 32 $\pm$ 1 & 6 $\pm$ 2.6 \\

       \hline
    \end{tabular}}
     \egroup

    \end{table}

% \begin{document}

%   \begin{center}
%     \begin{tabular}{ | c | c | c |}
%       \hline
%       \thead{A Head} & \thead{A Second \\ Head} & \thead{A Third \\ Head} \\
%       \hline
%       Some text &  \makecell{Some really \\ longer text}  & Text text text  \\
%       \hline
%     \end{tabular}
%   \end{center}

% \end{document}

\section{Experiments}
We evaluate three different algorithms using DriverGym to compare the effectiveness of these training strategies. The first one is an open-loop training baseline using L2 imitation loss (\textbf{SL}). Naive behavioral cloning is known to suffer from distribution shift between training and test data~\citep{Ross2011ARO}. We compare it with a stronger baseline, inspired by ChaufferNet~\citep{Bansal2019ChauffeurNetLT}, that alleviates distribution shift by introducing synthetic perturbations to the training trajectories (\textbf{SL + P}).

We also evaluate an RL policy, namely Proximal Policy Optimization (PPO)~\citep{Schulman2017ProximalPO} implemented in the SB3 framework~\citep{stable-baselines3}.  We choose PPO as it not only demonstrates remarkable performance but it is also empirically easy to tune~\citep{Schulman2017ProximalPO}. All the experiments have been performed on 2 Tesla T4 GPUs. The details of the model architectures, training strategies, hyperparameters used and experimental setup are provided in the appendix.

The performance of three runs (different seeds) of the three models on 100 real-world test scenes is reported in Table~\ref{compare}. Based on distance-based metrics, \textbf{PPO} is similar to \textbf{SL + P} in terms of ADE, however it suffers from high FDE. \textbf{PPO} showed fewer front and side collisions, however, it showed a much higher number of rear collisions, which can be explained by the passiveness of the ego vehicle. Finally, \textbf{SL} is the worst and corroborates the expectations.

% Performing an extensive comparison of RL algorithms is out of the scope of this work. All the experiments have been performed on 2 Tesla T4 GPU without extensive hyperparameter optimization for different RL algorithms.

% \section{Limitations}

% Another limitation of the current setup is the inferior baselines performance of reinforcement learning policies. We hope that DriverGym can democratize research in autonomous driving leading to the development of RL policies with improved performance on this challenging yet crucial task.

\section{Discussion}

We believe DriverGym is an important step towards solving the task of planning for autonomous driving. Thanks to its gym-compatible interface, it allows to easily train and evaluate RL policies for self-driving. Furthermore, surrounding agents can be controlled via a model trained on real-world data to improve their reactivity towards the ego. A current weakness of DriverGym is the time complexity of policy rollouts, which can be reduced through faster observation generation and mitigation of inter-process communication. 
% (see Appendix~\ref{policy_rollout} for more details).

% One weakness of DriverGym environment is the higher time complexity of policy rollouts when compared to game environments (i.e. Atari) due to the usage of real-world data and the rasterization process.

One avenue for future work is to provide fine-grained policy evaluation for different scene categories (e.g. restarting from an intersection controlled by a traffic light). We hope that DriverGym will provide a common ground for policy evaluation that is extensible, and will drive the improvement of the next generation of planning algorithms.

% we plan to extend DriverGym to support several different input representation, including inputs vectorization required to train an evaluate model such as VectorNet~\citep{Gao2020VectorNetEH}.

\section{Acknowledgements}
We would like to thank everyone at Level 5 working on data-driven planning, in particular Sergey
Zagoruyko, Alborz Alavian, Oliver Scheel, Yawei Ye, Moritz Niendorf, Stefano Pini and Maciej Wołczyk.

\bibliography{references}

% References follow the acknowledgments. Use unnumbered first-level heading for
% the references. Any choice of citation style is acceptable as long as you are
% consistent. It is permissible to reduce the font size to \verb+small+ (9 point)
% when listing the references.
% Note that the Reference section does not count towards the page limit.
% \medskip

% {
% \small

% [1] Alexander, J.A.\ \& Mozer, M.C.\ (1995) Template-based algorithms for
% connectionist rule extraction. In G.\ Tesauro, D.S.\ Touretzky and T.K.\ Leen
% (eds.), {\it Advances in Neural Information Processing Systems 7},
% pp.\ 609--616. Cambridge, MA: MIT Press.

% [2] Bower, J.M.\ \& Beeman, D.\ (1995) {\it The Book of GENESIS: Exploring
%   Realistic Neural Models with the GEneral NEural SImulation System.}  New York:
% TELOS/Springer--Verlag.

% [3] Hasselmo, M.E., Schnell, E.\ \& Barkai, E.\ (1995) Dynamics of learning and
% recall at excitatory recurrent synapses and cholinergic modulation in rat
% hippocampal region CA3. {\it Journal of Neuroscience} {\bf 15}(7):5249-5262.
% }

%%%%%%%%%%%%%%%%%%%%%%%%%%%%%%%%%%%%%%%%%%%%%%%%%%%%%%%%%%%%

\appendix

 \begin{minipage}{\textwidth}
  \begin{minipage}[b]{0.40\textwidth}
    \centering
    \includegraphics[width=0.92\textwidth]{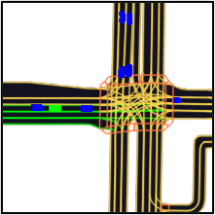}
    \\
    \includegraphics[width=0.92\textwidth]{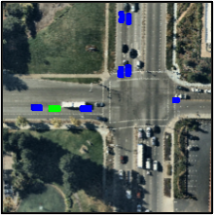}
    \captionof{figure}{Example Rasterization Modes}
    \label{raster_viz}
  \end{minipage}
  \hfill
  \begin{minipage}[b]{0.58\textwidth}
    \centering
    \includegraphics[width=\textwidth]{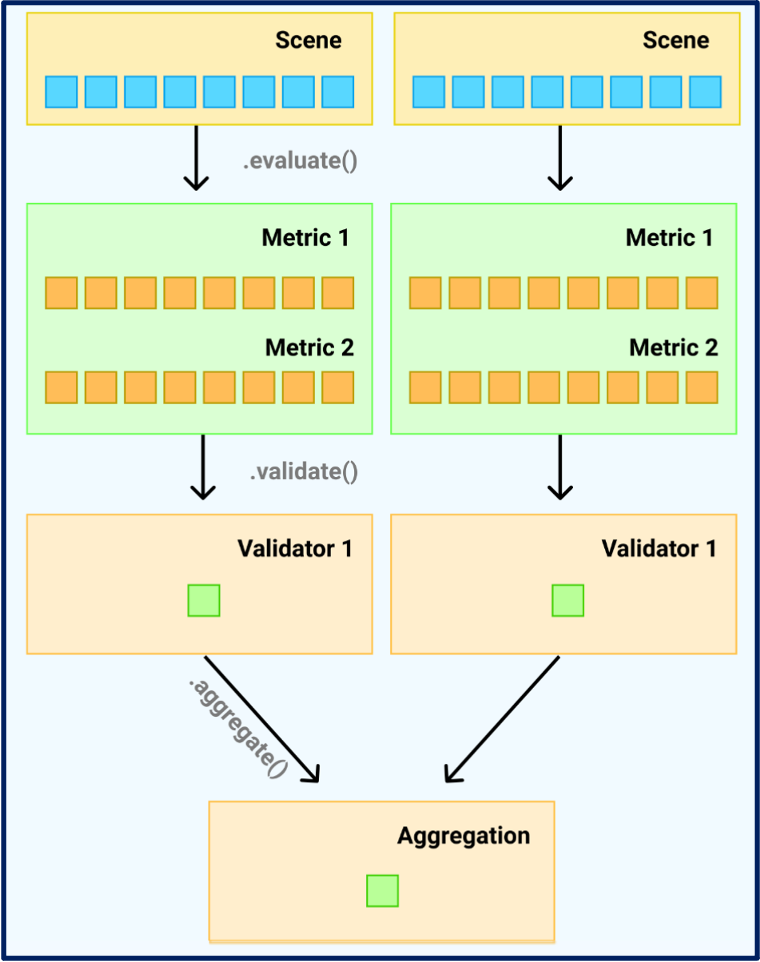}
    \captionof{figure}{Evaluation Plan}
    \label{eval_plan_viz}
    \end{minipage}
  \end{minipage}
 
\section{Appendix}

\subsection{Closed-Loop Evaluation Protocol}
\label{sec:cle}
Our CLE works on top of the simulation outputs provided by episode rollouts in DriverGym. An \textit{evaluation plan} (Fig.~\ref{eval_plan_viz}) is defined that comprises (1) metrics that are computed per frame (e.g. L2 displacement error) (2) validators that enforce constraints on the metrics per scene (L2 displacement error $\leq$ 4 meters), and (3) composite metrics per scene, that can depend both on the output of metrics and validators (e.g. passed driven miles). An example plan is provided in Listing~\ref{lst:cle-snippet}. Note that our CLE supports both reactive and non-reactive agents.

The evaluation protocol is flexible and new metrics can be easily incorporated to target specific cases of model failures. Within CLE, the user has access to all the simulation artifacts (trajectories, maps, \textit{log replay} data of ego and agents) while designing a new metric. We hope the DriverGym evaluation protocol facilitates researchers to diagnose targeted behaviors of their policies.

\begin{table}[]
    \centering
    \caption{Description of the various metrics provided in closed-loop evaluation protocol.}
    \label{metrics}
    \resizebox{\textwidth}{!}{\begin{tabular}{|c|c|c|}
        \hline
        Name & Type & Description \\
        \hline
        Average Displacement  &  Distance-Based &  \makecell{Computes the  L2 distance between predicted ego centroid and \\ ground-truth ego centroid averaged over the entire episode}\\ \hline
        Final Displacement  &  Distance-Based &  \makecell{Computes the L2 distance between predicted ego centroid and \\ ground-truth ego centroid at the last timestep of the episode}\\ \hline
        Distance to Reference  &  Distance-Based &  \makecell{Computes the L2 distance between the predicted centroid and the \\ closest waypoint in the reference trajectory (ground-truth ego)}\\ \hline
        Front Collision  &  Safety-Based &  \makecell{Computes whether a collision occurred between the front of ego and \\ any another agent}\\ \hline
        Side Collision  &  Safety-Based &  \makecell{ Computes whether a collision occurred between a side of ego and \\ any another agent}\\ \hline
        Rear Collision  &  Safety-Based &  \makecell{Computes whether a collision occurred between the rear of ego and \\ any another agent}\\
        \hline
    \end{tabular}}

\end{table}

\subsection{Model Architecture} \label{sec:backbone}

In our experiments, the backbone feature extractor is shared between the policy and the value networks. The feature extractor is composed of two convolutional networks followed by a fully connected layer, with ReLU activation. The feature extractor output is passed to both the policy and value networks composed of two fully connected layers with tanh activation. The open-loop baseline models have the same backbone architecture as above.

We perform group normalization after every convolutional layer. Empirically, we found that group normalization performs far superior to batch normalization. This can be attributed to the fact that activation statistics change quickly in on-policy algorithms (PPO is on-policy) while batch-norm learnable parameters can be slow to update causing training issues.

\subsection{Training}
The training data comprises $100$ scenes (average frame length $\sim 248$) where the initial frame is randomly sampled. The open-loop baseline models are trained in a supervised manner where the L2 loss is calculated on the predictions of 12 future time-steps (1.2 secs).

We train the PPO policy in closed-loop (the surrounding agents are \textit{log replayed}) for episodes of length $32$ time-steps. PPO policy network predicts the mean and standard deviation values of a gaussian to represent its actions. Further, the policy is initialized such that the initial actions are independent of the observations. Therefore, we normalize the action space (zero mean) for faster training convergence. For further training stability, we incorporate a unicycle kinematic model at the policy output, \textit{i.e.} the policy predicts the acceleration and steer.

For the PPO policy, we use an imitation loss-based reward. We define the reward as the negative of the L2 distance between the policy prediction and ego replay at every time-step. Reward clipping is performed for stability. Note that, DriverGym can also incorporate non-differentiable hand-crafted rules like collisions in the reward function to train different RL policies.

\subsection{Hyperparameters}
We train the PPO policy for $12M$ steps in which the learning rate is fixed to $3\mathrm{e}{-4}$ for the first $8M$ steps and then decreased by a factor of $10$ for the rest of the training. The discount factor is $0.80$ and GAE is $0.90$. $4$ environments are rolled out in parallel for a total of $1024$ time-steps before the model is updated for $10$ epochs on the collected rollout buffer. The mini-batch size of the model update is $64$ and the clipping parameter $\epsilon$ follows a linear decay schedule during training starting from $0.1$. The reward clipping threshold is fixed to $15$.

We train on  $112 \times 112$ pixel BEV rasters centered around the ego. The raster image is generated by combining the semantic map (3 channels) and the bounding boxes of the various agents in the scene (top image in Fig~\ref{raster_viz}). The past history and current bounding boxes of the agents (including ego) are incorporated via the channel dimension. We consider a history of 3 frames which along with the current frame leads to an additional 8 channels resulting in a raster image of size $112 \times 112 \times 11$.
The raster image is transformed such that it is centered around the ego vehicle.

\subsection{Visualizations}
The DriverGym environment provides the user with the ability to visualize the output of episode rollouts (see example in Fig~\ref{fig:rollout}). The visualization is carried out using the Bokeh interaction visualization library~\citep{Bokeh}.

\subsection{DriverGym API Snippets}

% Furthermore we quantitatively benchmark the different steps of the policy rollout in Tab~\ref{tab:benchmark}. 

% \begin{table}[]
%     \centering
%     \begin{tabular}{c|c}
%         Name & Time (Mins) \\
%         DriverGym Policy Simulation &  \\
%         Environment Update & \\
%         Policy Forward Pass & \\
%         Inter-process communication & \\
%     \end{tabular}
%     \caption{Caption}
%     \label{tab:benchmark}
% \end{table}

\begin{listing}[H]
\begin{minted}[mathescape,
               linenos,
               numbersep=5pt,
               frame=lines,
               framesep=2mm]{python}

from stable_baselines3 import PPO
import gym

env = gym.make("drivergym-v0")
model = PPO("CnnPolicy", env)
model.learn(n_steps=1000000)
\end{minted}
\caption{Code snippet showing the user API for using DriverGym environment with Stable Baselines3~\citep{stable-baselines3}.}
\label{lst:cle-snippet}
\end{listing}

\begin{listing}[H]
\begin{minted}[mathescape,
              linenos,
              numbersep=5pt,
              frame=lines,
              framesep=2mm]{python}
              
from l5kit.cle.metrics import SupportsMetricCompute

class L2DisplacementErrorMetric(SupportsMetricCompute):
    metric_name = "l2_displacement_error"

    def compute(self, simulation_output: SimulationOutputCLE) -> torch.Tensor:
        simulated_scene_ego_state = simulation_output.simulated_ego_states
        simulated_centroid = simulated_scene_ego_state[:, :2]
        observed_ego_states = simulation_output.recorded_ego_states[:, :2]
        return torch.norm(simulated_centroid - observed_ego_states_fraction)

\end{minted}
\caption{Code snippet showing the flexibility of adding new metrics to closed-loop evaluator (CLE). In this example, we are defining a L2 displacement error metric.}
\label{lst:cle-snippet}
\end{listing}

\end{document}